\documentclass[letterpaper, 10 pt, conference]{IEEEtran}

\IEEEoverridecommandlockouts

\usepackage{amsmath,amssymb,amsfonts}
\usepackage{algorithmic}
\usepackage{graphicx}
\usepackage{textcomp}
\usepackage{xcolor}
\usepackage{varwidth}
\usepackage{subfigure}
\usepackage{cite}
\usepackage{comment}
\usepackage{balance}

\begin{document}

\title{Towards Intuitive Drone Operation Using a Handheld Motion Controller}

\author{\IEEEauthorblockN{Daria Trinitatova}
\IEEEauthorblockA{\textit{Center for Digital Engineering} \\
\textit{Skoltech }\\
Moscow, Russia \\
daria.trinitatova@skoltech.ru}
\and
\IEEEauthorblockN{Sofia Shevelo}
\IEEEauthorblockA{\textit{Center for Digital Engineering} \\
\textit{Skoltech }\\
Moscow, Russia \\
sofia.shevelo@skoltech.ru}
\and
\IEEEauthorblockN{Dzmitry Tsetserukou}
\IEEEauthorblockA{\textit{Center for Digital Engineering} \\
\textit{Skoltech }\\
Moscow, Russia \\
d.tsetserukou@skoltech.ru}

}

\maketitle

\begin{abstract}

We present an intuitive human-drone interaction system that utilizes a gesture-based motion controller to enhance the drone operation experience in real and simulated environments. The handheld motion controller enables natural control of the drone through the movements of the operator's hand, thumb, and index finger: the trigger press manages the throttle, the tilt of the hand adjusts pitch and roll, and the thumbstick controls yaw rotation. Communication with drones is facilitated via the ExpressLRS radio protocol, ensuring robust connectivity across various frequencies. The user evaluation of the flight experience with the designed drone controller using the UEQ-S survey showed high scores for both Pragmatic (mean=2.2, SD = 0.8) and Hedonic (mean=2.3, SD = 0.9) Qualities. This versatile control interface supports applications such as research, drone racing, and training programs in real and simulated environments, thereby contributing to advances in the field of human-drone interaction. 
 
\end{abstract}

\section{Introduction}


In recent years, there has been significant progress in Human-Robot Interaction (HRI), particularly in the development of various interaction interfaces to facilitate drone control.
One of the common approaches includes the design of a physical interface for robot operation with a set of motion capture (mocap) sensors and/or electromyography (EMG) sensors. Thus, in \cite{rognon2018flyjacket}, a wearable interface based on a soft exoskeleton was developed to control the UAV using gestures of the upper body. It includes a motion-tracking device to monitor body movements and an arm support system to prevent fatigue. In addition, the control interface is paired with a virtual reality (VR) headset to provide first-person view (FPV) from the UAV.  
In \cite{gromov2020guiding}, a human-robot interface was introduced for precise landing of a quadrotor based on pointing gestures. The designed system consists of a pair of commercial Myo armbands placed on the upper arm and forearm to detect the pointed direction of the operator. The application of a vibrotactile glove integrated with a mocap system for collision avoidance during drone teleoperation was presented in \cite{macchini2020hand}. The vibrotactile feedback was used to provide the information about the quadcopter movement along the up-down, left-right, front-back directions to augment the user's awareness of the quadcopter state, especially in the case of limited visibility conditions. Similarly, in \cite{ramachandran2021arm}, a wearable haptic sleeve was proposed that provides kinesthetic feedback to the operator to avoid obstacles during teleoperation, allowing natural drone movement when visual feedback is inactive. Expanding drone control using a wearable device detected arm stiffening and fist clenching from EMG signals and rotation gestures from IMU signals was introduced in \cite{delpreto2020plug}. In further processing, the arm gesture data obtained are transformed into drone control commands.

\begin{figure}[!t]
    \centering
    \includegraphics[width=0.92\linewidth]{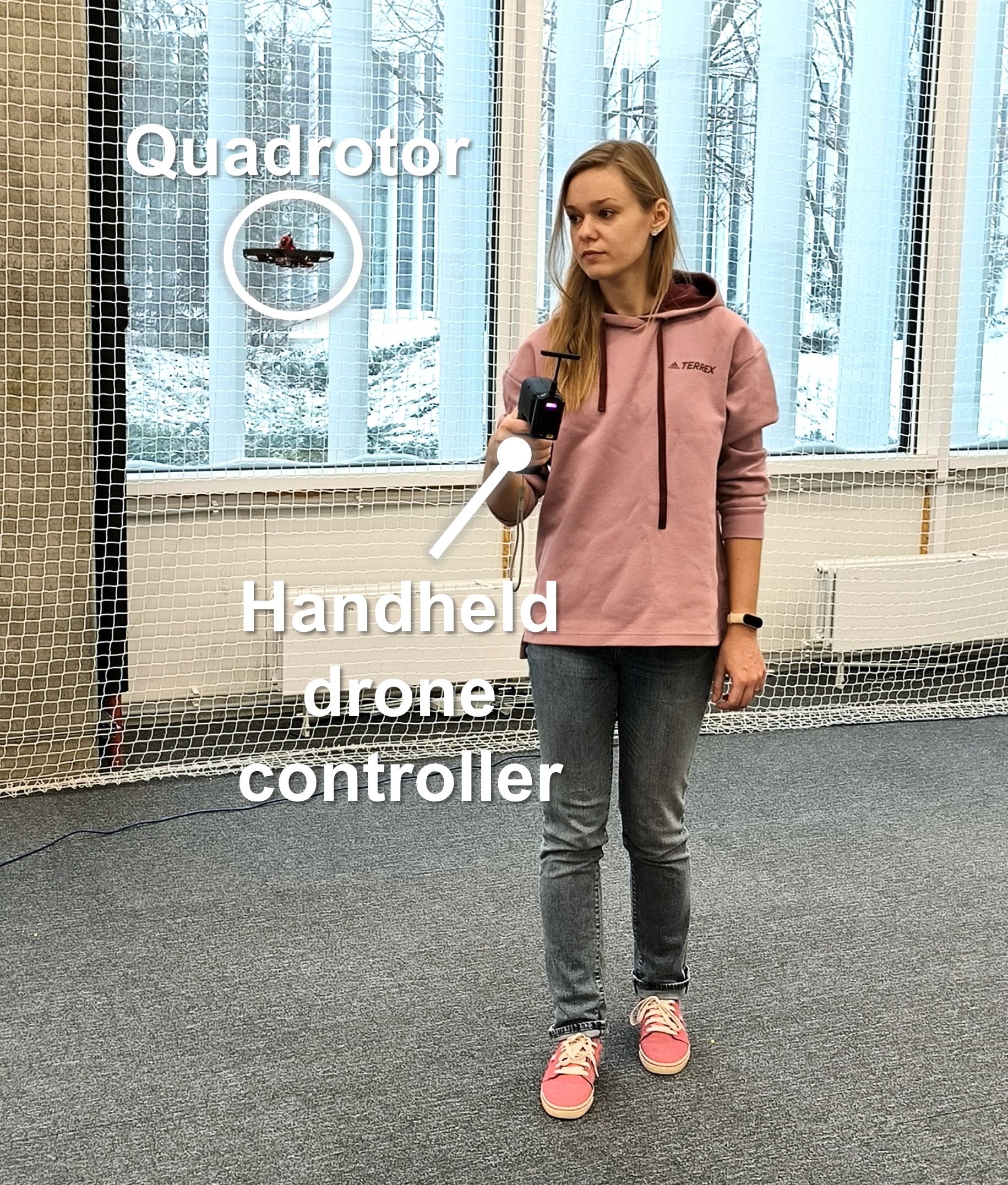}
    \caption{An operator during the drone piloting using a developed motion controller.}
    \label{fig:mr}
\end{figure}

AI-driven interfaces have also been explored to facilitate drone control through the recognition of voice commands and body or hand gestures \cite{menshchikov2019data}. In \cite{miehlbradt2018data}, optimal body movement patterns were examined using a body-machine interface for drone control, reducing the need for extensive sensor coverage. By analyzing upper-body kinematics and muscle activity, the study enhanced accuracy in drone control, allowing users to intuitively steer the drone through simplified body movements. A wearable drone controller with vibrotactile feedback was presented in \cite{lee2023wearable}. The quadcopter is controlled using hand gestures sensed by the IMU and classified using machine learning, and a vibration actuator is used to alert the operator when approaching an obstacle. The application of DNN-based gesture control for drone racing was introduced in \cite{serpiva2024omnirace}. Hand-gesture recognition was implemented through a web camera placed in front of the operator. In \cite{abdi2023safe}, a human-swarm interface was presented for safe operation using gesture-based control and haptic feedback. An EMG-based wristband is used to control swarm formation using a trained onboard AI model to recognize eight user gestures, and a haptic vest enhances the operator's spatial awareness of swarm location using vibrotactile feedback.

In addition to the development of physical control interfaces, the application of extended reality technologies to visual interfaces for UAV operators can create assistive systems that enhance operational efficiency and safety of drone operations. For example, a system for natural human-drone interaction (HDI) using a projected graphical user interface and gestural control was proposed in \cite{cauchard2019drone}. In \cite{yashin2019aerovr}, the application of VR environment to the design of a teleoperation system for aerial manipulation was explored. The proposed system allows operating a UAV equipped with a 4-DoF robotic arm and embedded sensors using a wearable interface with vibrotactile feedback using a digital twin, enabling precise control and interaction during teleoperated tasks. In \cite{chen2021pinpointfly}, a mobile augmented reality (AR) interface was designed for HDI, providing operators with the ability to control the position and rotation of the drone. The AR-based interface provides egocentric view and FPV for drone control, augmented with a virtual drone and the shadow cast by the drone to improve the operator’s spatial perception. In \cite{sautenkov2024flightar}, an AR interface was proposed to assist UAV operators. It includes AR overlays of different camera feeds from the UAV on the operator's view, augmented by object detection. The application of mixed reality (MR) for teleoperation of aerial robots was presented in \cite{allenspach2023design}. Using a holographic user interface, an operator can control a separate translational or rotational axis for precise movement or manipulate all DoFs of the aerial vehicle simultaneously. 

The growing diversity of drone applications highlights the need for a versatile control interface compatible with UAVs of different frame sizes and types, including research and education platforms, as well as FPV racing drones. Conventional commercial remote controls usually require holding them in both hands and controlling the drone's main inputs with thumbsticks, which is often not intuitive for novice users. Another challenge is that many modern controllers often rely on proprietary communication protocols, limiting cross-platform compatibility. For example, DJI company has proposed a handheld controller for intuitive drone control, but with a proprietary protocol compatible only with DJI drones.  An open-source communication protocol, capable of connecting with various types of drones, would enhance interoperability and innovation across fields. 

In this work, we present a novel human-drone interaction interface in the form of a handheld motion controller designed to enhance the drone operation experience. The developed motion controller can be easily integrated with drones employing various flight controller software via the ExpressLRS (ELRS) open-source radio protocol. In addition, it is compatible with simulated environments for operator training through Bluetooth communication. The controller design allows users to intuitively manipulate drone movements: the trigger manages throttle, hand tilt adjusts pitch and roll angles, and the thumbstick controls yaw rotation. 

\begin{figure}[!t]
    \centering
    \includegraphics[width=0.95\linewidth]{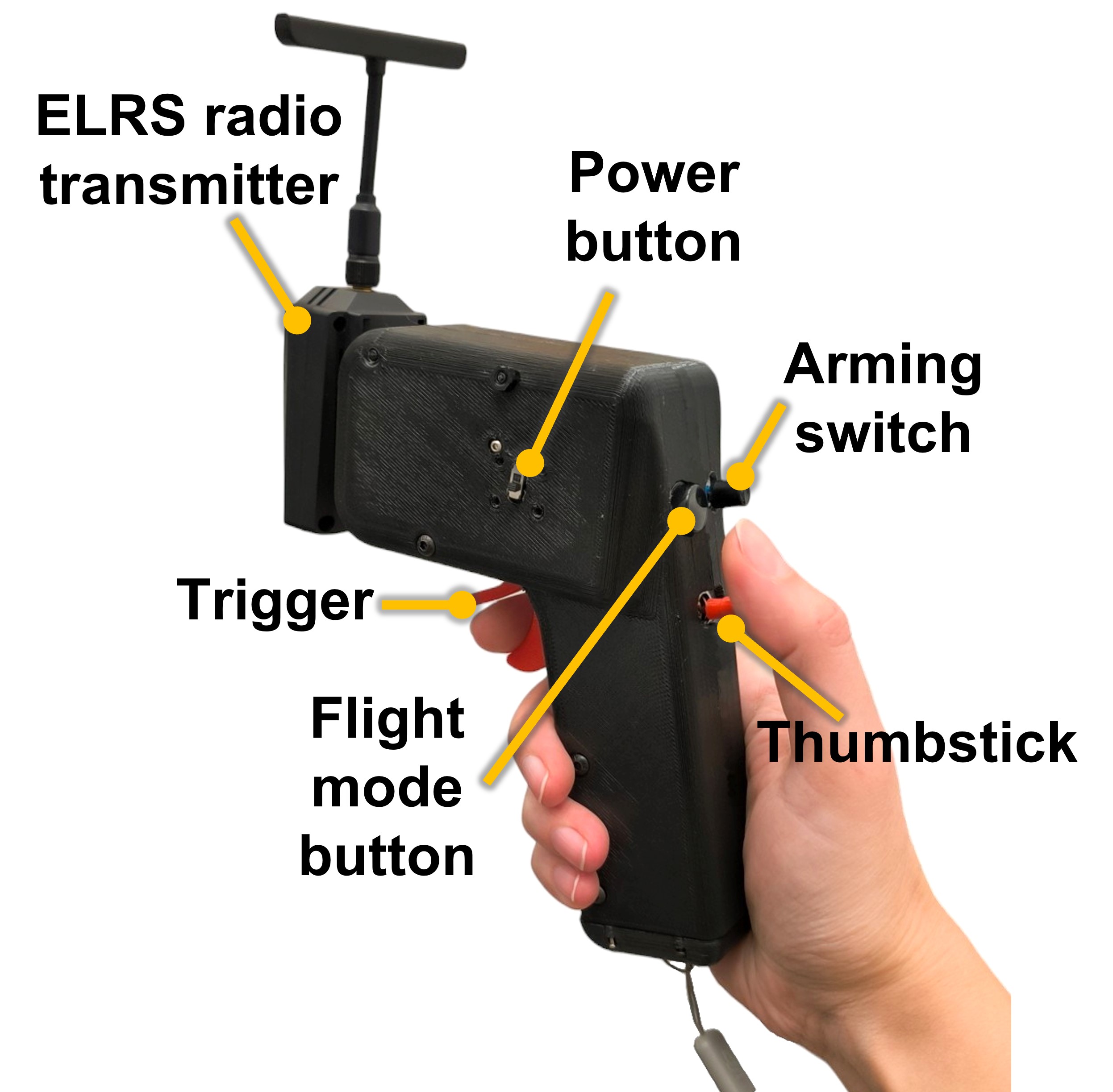}
    \caption{A prototype of the developed handheld drone controller.}
    \label{fig:prototype}
\end{figure}
\section{System Overview}

\subsection{Handheld Motion Controller}
\begin{figure}[!t]
  \centering
  \subfigure[Pitch control.]{\includegraphics[width=0.5\linewidth]{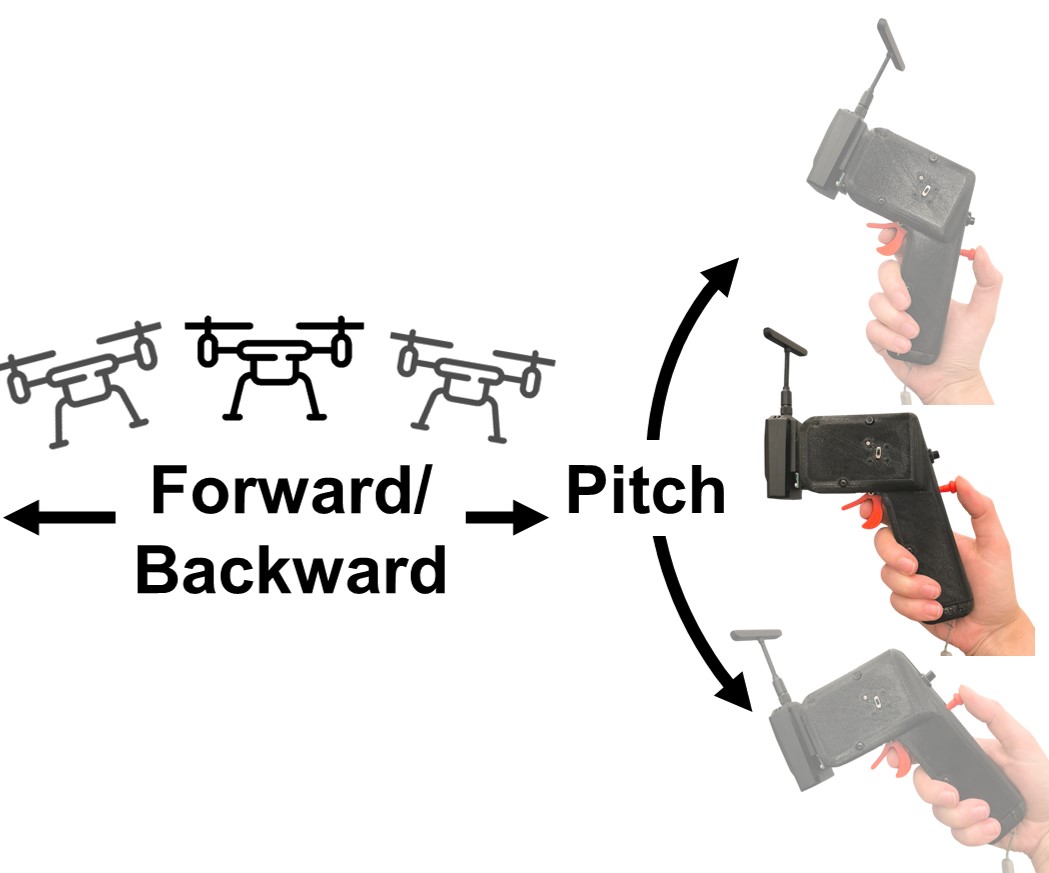}}
   \subfigure[Roll control.]{\includegraphics[width=0.4\linewidth]{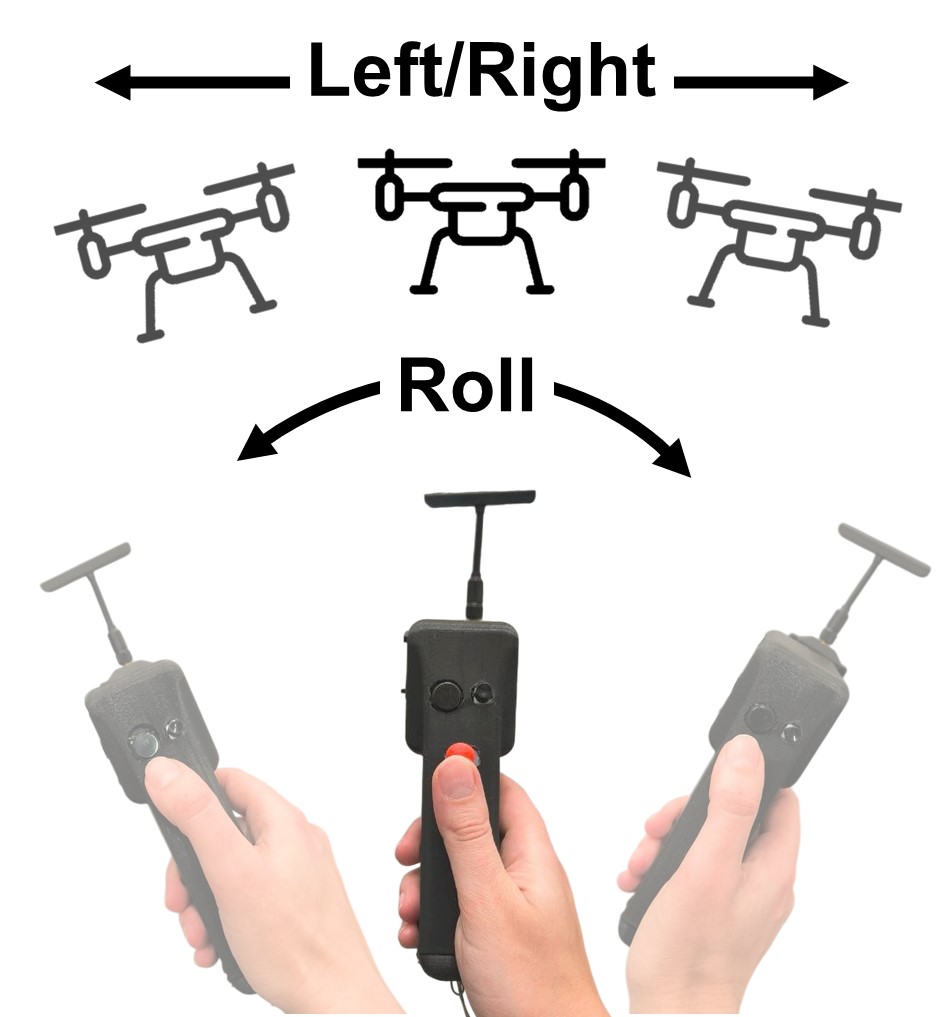}}
   \subfigure[Throttle control.]{\includegraphics[width=0.46\linewidth]{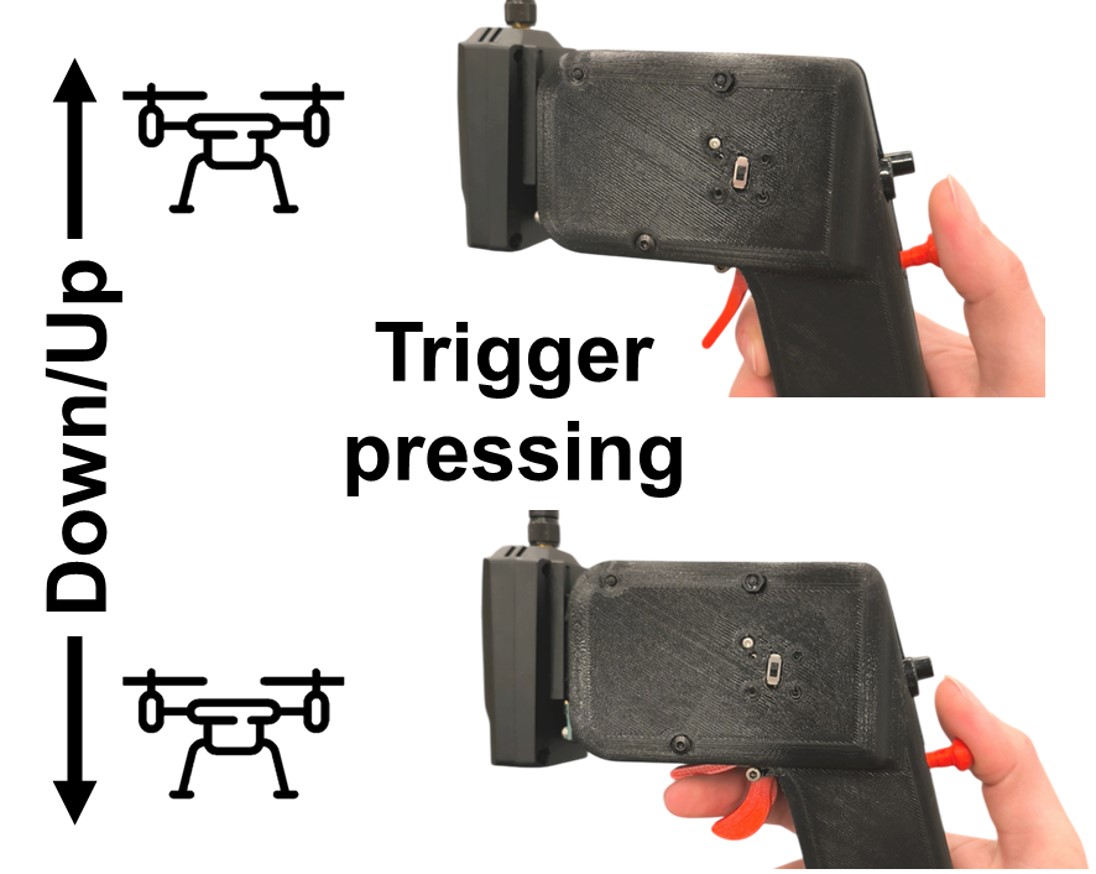}}
   \subfigure[Yaw control.]{\includegraphics[width=0.44\linewidth]{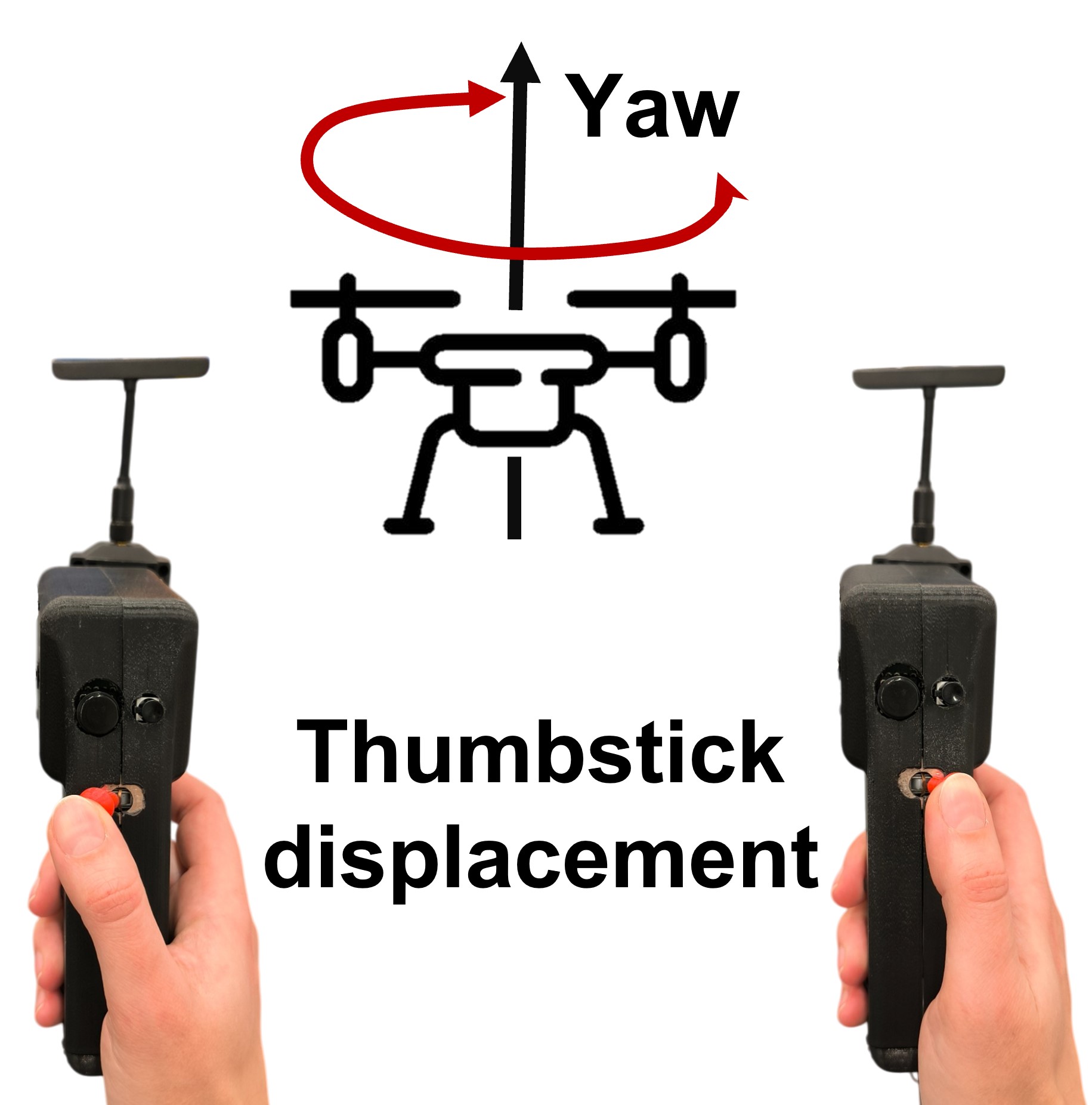}}
  \caption{An illustration of the drone control principle.}\label{fig:control}
\end{figure}

The designed motion controller is aimed at intuitive drone control using movements of one hand (Fig. \ref{fig:prototype}). The prototype components were printed from PLA material. The motion controller is built based on the ESP32 microcontroller, dual-axis tilt sensor, and interactive buttons. The designed controller allows direct control of 4 input channels for UAV flight. The operating principle is as follows: the trigger manages the drone’s throttle  (Fig. \ref{fig:control} (c)), the controller's tilts along the pitch and roll axes are converted to the pitch and roll control inputs responsible for the linear movement of the drone (Fig. \ref{fig:control} (a)-(b)) and the thumbstick is responsible for rotating the drone around the yaw axis (Fig. \ref{fig:control} (d)). Communication with the drone is carried out through the ELRS radio protocol, ensuring reliable and low-latency connectivity. ELRS supports various operating frequencies, thereby the controller is compatible with different external ELRS transmitters, expanding the diversity of drones for piloting. In addition, ESP32 microcontroller supports Bluetooth, which allows a simple communication channel with different simulated environments, VR and MR applications.

\subsection{Drone Control Use Cases}
\subsubsection{Training of drone piloting skills in real conditions} Utilizing ELRS communication, the developed motion controller can be easily integrated with popular FPV models, such as Tiny Whoop microdrones, CineWhoop cinematic shooting drones. 
The designed drone controller facilitates novice users to train piloting skills in indoor and outdoor environments. For drone racing application, the controller can be used with FPV goggles to practice maneuvering the drone with different flight modes in obstacle-ridden environments. 

\subsubsection{Simulated environment for training}
The developed controller can also be connected to commercial drone racing simulators such as LiftOff and VelociDrone for initial pilot training and familiarity with drone operations, or to open-source drone simulators for testing drone control algorithms via USB serial port or Bluetooth. 

In addition, we designed a mixed reality (MR) application to support operator training in empty spaces with virtual models of the training environment (Fig. \ref{fig:mr}).
The designed MR application allows creating and seamlessly integrating a virtual training area with real user's space, as well as preliminary piloting training using a virtual model of the drone. Utilizing Unity engine and Meta Quest 3 head-mounted display (HMD), we seamlessly integrate virtual components such as drone gates, obstacles, landing pads with real-world environments.
Upon entering the application, the user wearing the HMD scans the surfaces of the room, capturing environmental surfaces. With the use of a VR controller or hand tracking feature, users can freely manipulate, scale, and delete virtual objects within the environment. 
The seamless integration of physical space with virtual objects allows users to practice piloting skills in any uncluttered environment.

\begin{figure}
    \centering
{\includegraphics[width=0.98\linewidth]{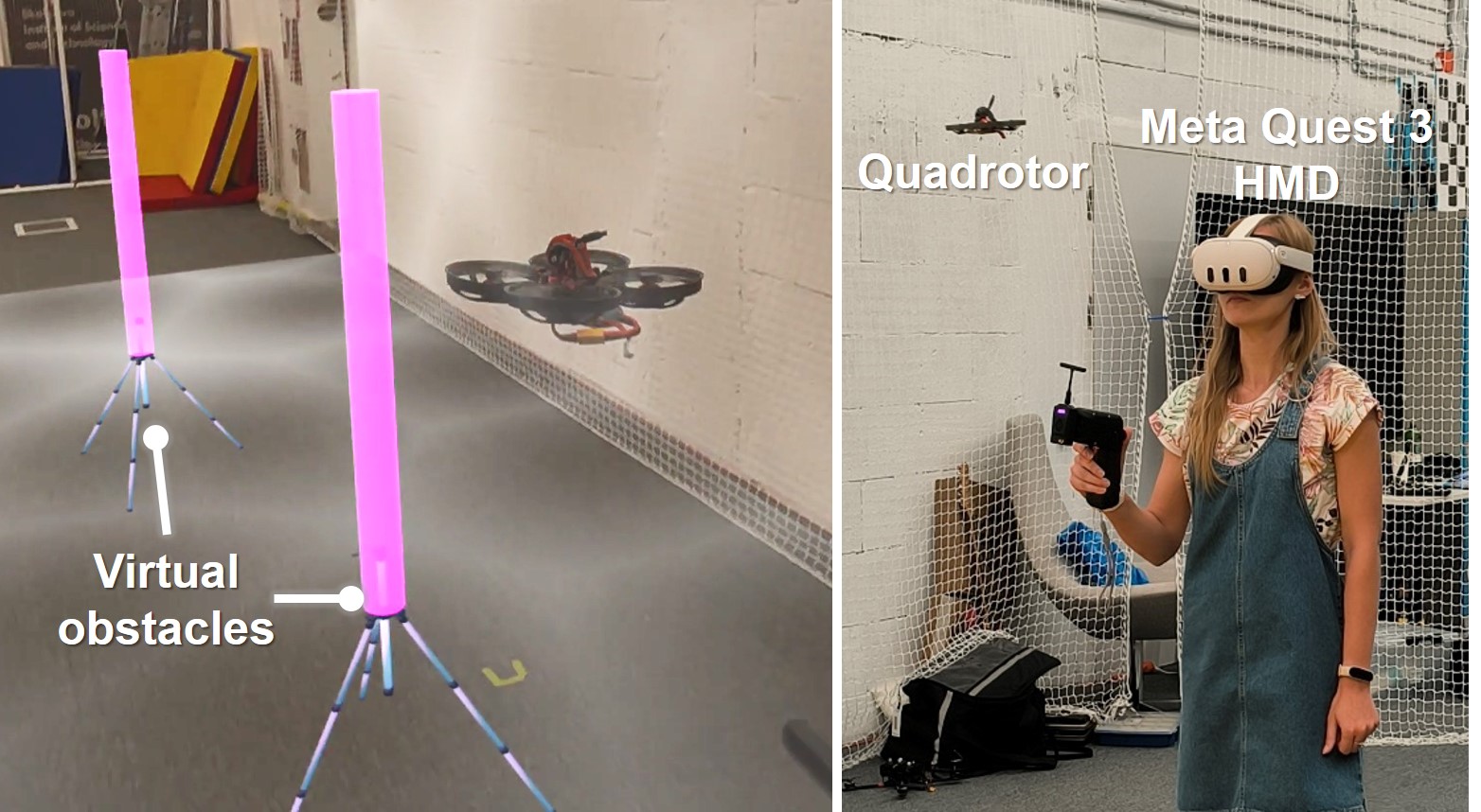}}
    \caption{Drone control through MR application: operator's view from the HMD (left), operator wearing the HMD and with handheld controller (right).}
    \label{fig:mr}
\end{figure}

\section{User Study}

\subsection{Experimental Description}
We conducted a preliminary experiment to evaluate the intuitiveness of drone control with the developed motion controller during short flight sessions with training exercises. 
Ten participants (3 females), aged 23–35 years (mean = 26.3, SD = 3.4), with no prior or limited experience with drone control, volunteered to participate in the experiment, giving their informed consent. 
\begin{figure}[!h]
  \centering
  \subfigure[]{\includegraphics[width=0.95\linewidth]{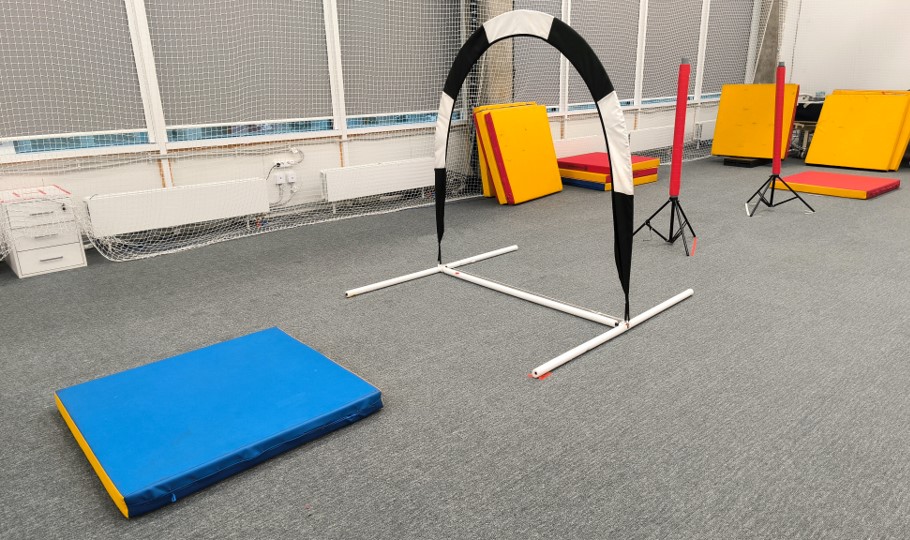}}
   \subfigure[]{\includegraphics[width=0.95\linewidth]{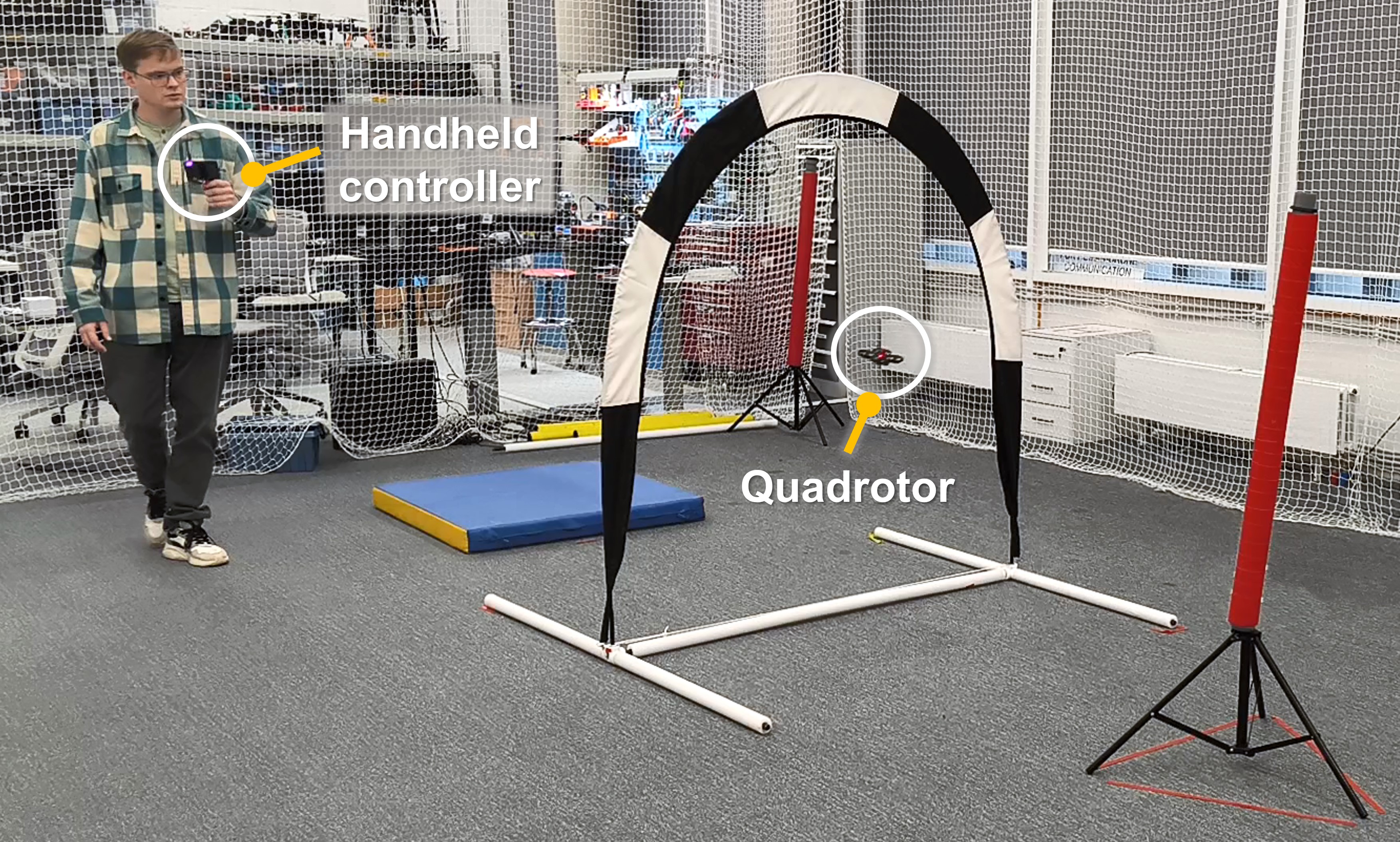}}
  \caption{(a) Flight area with a playground for drone control exercises for the experiment. (b) A participant during the drone piloting experiment.}\label{fig:us}
\end{figure}

The experimental procedure comprised two steps. Firstly, the participants were instructed on basic principles of drone control, after which they practiced controlling the basic movements of the drone during flight such as altitude changing, linear movement (forward/backward, right/left), clockwise and counterclockwise rotation of the drone. After this training, participants were asked to complete a flight task along the track. The flight area for the experiment consisted of two zones, one of which was free space to practice basic skills during the first step, and the second consisted of a simplified track with a gate and a pair of vertical obstacles (Fig. \ref{fig:us} (a)). While flying along the track, participants could follow the drone to maintain a view of the drone throughout the track (Fig. \ref{fig:us} (b)). A Happymodel Mobula 8 microdrone with Angle flight mode was used for the experiment. Each experimental session lasted from 5 to 10 minutes.
After the flight session, we asked participants to evaluate their experience of using the designed handheld drone controller using a Short-User Experience Questionnaire (UEQ-S) \cite{ueq-s}, comprised of eight items.

\subsection{Experimental Results}
\begin{figure}[b]
\centering
\includegraphics[width=0.98\linewidth]{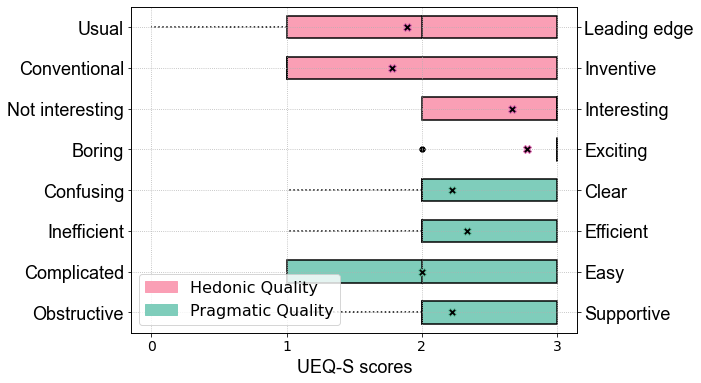} 
\caption{UEQ-S results in a scale range of -3 to 3. Crosses mark average values.}
\label{fig:ueq}
\end{figure}

The results of the Short-User Experience Questionnaire are shown in Fig. \ref{fig:ueq}. The first four scores represent the Pragmatic Quality scale and the last four items the Hedonic Quality scale. Overall, participants evaluated all the UEQ-S items positively ($\mu>0.8$). 

The highest scores for user experience with the drone controller were obtained for the Hedonic Quality scale (the scores obtained for Interest and Excitement items comprised $\mu=2.7$, $SD = 0.5$ and $\mu=2.8$, $SD = 0.4$). The averaged results for obtained UEQ-S scales are shown in Table \ref{tab1}. All participants rated their experience highly both in terms of Pragmatic ($\mu=2.2$, $SD = 0.8$) and Hedonic ($\mu=2.3$, $SD = 0.9$) Qualities. These results confirm the potential of the developed controller to facilitate drone piloting training for novice users.
\begin{table}[h]
\caption{UEQ-S Scales}
\begin{center}
\begin{tabular}{|c|c|c|}
\hline
\textbf{Pragmatic Quality}& \textbf{Hedonic Quality}& \textbf{Overall} \\
\hline
 2.2 $\pm$ 0.8 & 2.3 $\pm$ 0.9 & 2.2 $\pm$ 0.8 \\
\hline
\end{tabular}
\label{tab1}
\end{center}
\end{table}

\section{Conclusion}

We have proposed a handheld control interface for natural drone piloting using the movements of the user's hand. The presented system enhances user interaction through intuitive movement alignment of user's hand with corresponding drone input signals, facilitating applications in drone racing, education, research, etc. The experimental evaluation revealed high user evaluation of both Pragmatic ($\mu=2.2$) and Hedonic Qualities ($\mu=2.3$) during flight experience with the developed drone controller. These results indicate the potential of the proposed control interface to lower the entry threshold and improve the UAV piloting experience. 

Future directions include improving the ergonomics of the drone controller, conducting extensive user evaluation in operating the drone using the proposed controller, and expanding the functionality of the MR platform for immersive drone piloting. 
 
\section{ACKNOWLEDGMENT} 
Research reported in this publication was financially supported by the RSF grant No. 24-41-02039.

\bibliographystyle{IEEEtran}
\balance
\bibliography{bibi}

\begin{thebibliography}{10}
\providecommand{\url}[1]{#1}
\csname url@samestyle\endcsname
\providecommand{\newblock}{\relax}
\providecommand{\bibinfo}[2]{#2}
\providecommand{\BIBentrySTDinterwordspacing}{\spaceskip=0pt\relax}
\providecommand{\BIBentryALTinterwordstretchfactor}{4}
\providecommand{\BIBentryALTinterwordspacing}{\spaceskip=\fontdimen2\font plus
\BIBentryALTinterwordstretchfactor\fontdimen3\font minus \fontdimen4\font\relax}
\providecommand{\BIBforeignlanguage}[2]{{%
\expandafter\ifx\csname l@#1\endcsname\relax
\typeout{** WARNING: IEEEtran.bst: No hyphenation pattern has been}%
\typeout{** loaded for the language `#1'. Using the pattern for}%
\typeout{** the default language instead.}%
\else
\language=\csname l@#1\endcsname
\fi
#2}}
\providecommand{\BIBdecl}{\relax}
\BIBdecl

\bibitem{rognon2018flyjacket}
C.~Rognon, S.~Mintchev, F.~Dell'Agnola, A.~Cherpillod, D.~Atienza, and D.~Floreano, ``Flyjacket: An upper body soft exoskeleton for immersive drone control,'' \emph{IEEE Robotics and Automation Letters}, vol.~3, no.~3, pp. 2362--2369, 2018.

\bibitem{gromov2020guiding}
B.~Gromov, L.~Gambardella, and A.~Giusti, ``Guiding quadrotor landing with pointing gestures,'' in \emph{Human-Friendly Robotics 2019: 12th International Workshop}.\hskip 1em plus 0.5em minus 0.4em\relax Springer, 2020, pp. 1--14.

\bibitem{macchini2020hand}
M.~Macchini, T.~Havy, A.~Weber, F.~Schiano, and D.~Floreano, ``Hand-worn haptic interface for drone teleoperation,'' in \emph{2020 IEEE International Conference on Robotics and Automation (ICRA)}, 2020, pp. 10\,212--10\,218.

\bibitem{ramachandran2021arm}
V.~Ramachandran, M.~Macchini, and D.~Floreano, ``Arm-wrist haptic sleeve for drone teleoperation,'' \emph{IEEE Robotics and Automation Letters}, vol.~7, no.~4, pp. 12\,054--12\,061, 2022.

\bibitem{delpreto2020plug}
J.~DelPreto and D.~Rus, ``Plug-and-play gesture control using muscle and motion sensors,'' in \emph{Proceedings of the 2020 ACM/IEEE International Conference on Human-Robot Interaction}, 2020, pp. 439--448.

\bibitem{menshchikov2019data}
A.~Menshchikov, D.~Ermilov, I.~Dranitsky, L.~Kupchenko, M.~Panov, M.~Fedorov, and A.~Somov, ``Data-driven body-machine interface for drone intuitive control through voice and gestures,'' in \emph{IECON 2019-45th Annual Conference of the IEEE Industrial Electronics Society}, vol.~1, 2019, pp. 5602--5609.

\bibitem{miehlbradt2018data}
J.~Miehlbradt, A.~Cherpillod, S.~Mintchev, M.~Coscia, F.~Artoni, D.~Floreano, and S.~Micera, ``Data-driven body--machine interface for the accurate control of drones,'' \emph{Proceedings of the National Academy of Sciences}, vol. 115, no.~31, pp. 7913--7918, 2018.

\bibitem{lee2023wearable}
J.-W. Lee and K.-H. Yu, ``Wearable drone controller: Machine learning-based hand gesture recognition and vibrotactile feedback,'' \emph{Sensors}, vol.~23, no.~5, p. 2666, 2023.

\bibitem{serpiva2024omnirace}
V.~Serpiva, A.~Fedoseev, S.~Karaf, A.~A. Abdulkarim, and D.~Tsetserukou, ``Omnirace: 6d hand pose estimation for intuitive guidance of racing drone,'' in \emph{2024 IEEE/RSJ International Conference on Intelligent Robots and Systems (IROS)}, 2024, pp. 2508--2513.

\bibitem{abdi2023safe}
S.~S. Abdi and D.~A. Paley, ``Safe operations of an aerial swarm via a cobot human swarm interface,'' in \emph{2023 IEEE International Conference on Robotics and Automation (ICRA)}, 2023, pp. 1701--1707.

\bibitem{cauchard2019drone}
J.~R. Cauchard, A.~Tamkin, C.~Y. Wang, L.~Vink, M.~Park, T.~Fang, and J.~A. Landay, ``Drone.io: A gestural and visual interface for human-drone interaction,'' in \emph{2019 14th ACM/IEEE International Conference on Human-Robot Interaction (HRI)}, 2019, pp. 153--162.

\bibitem{yashin2019aerovr}
G.~A. Yashin, D.~Trinitatova, R.~T. Agishev, R.~Ibrahimov, and D.~Tsetserukou, ``Aero{VR}: Virtual reality-based teleoperation with tactile feedback for aerial manipulation,'' in \emph{2019 19th International Conference on Advanced Robotics (ICAR)}, 2019, pp. 767--772.

\bibitem{chen2021pinpointfly}
L.~Chen, K.~Takashima, K.~Fujita, and Y.~Kitamura, ``Pinpointfly: An egocentric position-control drone interface using mobile {AR},'' in \emph{Proceedings of the 2021 CHI Conference on Human Factors in Computing Systems}, 2021, pp. 1--13.

\bibitem{sautenkov2024flightar}
O.~Sautenkov, S.~Asfaw, Y.~Yaqoot, M.~A. Mustafa, A.~Fedoseev, D.~Trinitatova, and D.~Tsetserukou, ``Flight{AR}: {AR} flight assistance interface with multiple video streams and object detection aimed at immersive drone control,'' in \emph{2024 IEEE International Conference on Robotics and Biomimetics (ROBIO)}, 2024, pp. 614--619.

\bibitem{allenspach2023design}
M.~Allenspach, T.~K{\"o}tter, R.~B{\"a}hnemann, M.~Tognon, and R.~Siegwart, ``Design and evaluation of a mixed reality-based human-robot interface for teleoperation of omnidirectional aerial vehicles,'' in \emph{2023 International Conference on Unmanned Aircraft Systems (ICUAS)}, 2023, pp. 1168--1174.

\bibitem{ueq-s}
M.~Schrepp, A.~Hinderks, and J.~Thomaschewski, ``Design and evaluation of a short version of the user experience questionnaire (ueq-s),'' \emph{International Journal of Interactive Multimedia and Artificial Intelligence, 4 (6), 103-108.}, 2017.

\end{thebibliography}

\end{document}